\documentclass[10pt,twocolumn,letterpaper]{article}
\usepackage{cvpr}
\usepackage{times}
\usepackage{epsfig}
\usepackage{graphicx}
\usepackage{amsmath}
\usepackage{amssymb}

\usepackage{graphicx}
\usepackage{multirow}
\usepackage{floatrow}
\usepackage{animate}
\usepackage{subfig}
\newfloatcommand{capbtabbox}{table}[][\FBwidth]
\usepackage[misc]{ifsym}
\usepackage{pdfpages}

\cvprfinalcopy % *** Uncomment this line for the final submission

\def\cvprPaperID{727} % *** Enter the CVPR Paper ID here

\ifcvprfinal\fi
\begin{document}

%%%%%%%%% TITLE
%\title{Video Anomaly Detection as Learning from Noisy Labels:\\Leverage the Power of Action Classifiers}
\title{Graph Convolutional Label Noise Cleaner:\\Train a Plug-and-play Action Classifier for Anomaly Detection}
% For a paper whose authors are all at the same institution,
% omit the following lines up until the closing ``}''.
% Additional authors and addresses can be added with ``\and'',
% just like the second author.
% To save space, use either the email address or home page, not both
\author{Jia-Xing Zhong$^{1,2}$~~~Nannan Li$^{3,1,2}$~~~Weijie Kong$^{1,2}$~~~Shan Liu$^{4}$~~~Thomas H. Li$^{1}$~~~{Ge Li \footnotesize{\Letter}}$^{1,2}$\\
	$^1$School of Electronic and Computer Engineering, Peking University~~~$^2$Peng Cheng Laboratory~~~\\
	$^3$Institute of Intelligent Video Audio Technology, Longgang Shenzhen~~~~$^4$Tencent America\\
	{\tt\small jxzhong@pku.edu.cn~~~lnnsiat@gmail.com~~~weijie.kong@pku.edu.cn}\\ 	{\tt\small shanl@tencent.com~~~~~~tli@aiit.org.cn~~~~~~geli@ece.pku.edu.cn}
}

\maketitle
%\thispagestyle{empty}

%%%%%%%%% ABSTRACT
\begin{abstract}
Video anomaly detection under weak labels is formulated as a typical multiple-instance learning problem in previous works. In this paper, we provide a new perspective, \ie, a supervised learning task under noisy labels. In such a viewpoint, as long as cleaning away label noise, we can directly apply fully supervised action classifiers to weakly supervised anomaly detection, and take maximum advantage of these well-developed classifiers. For this purpose, we devise a graph convolutional network to correct noisy labels. Based upon feature similarity and temporal consistency, our network propagates supervisory signals from high-confidence snippets to low-confidence ones. In this manner, the network is capable of providing cleaned supervision for action classifiers. During the test phase, we only need to obtain snippet-wise predictions from the action classifier without any extra post-processing. Extensive experiments on 3 datasets at different scales with 2 types of action classifiers demonstrate the efficacy of our method. Remarkably, we obtain the frame-level AUC score of 82.12\% on UCF-Crime.
\end{abstract}

%%%%%%%%% BODY TEXT
\section{Introduction}
\begin{figure}[ht]
\centering
\includegraphics[width=0.9\textwidth]{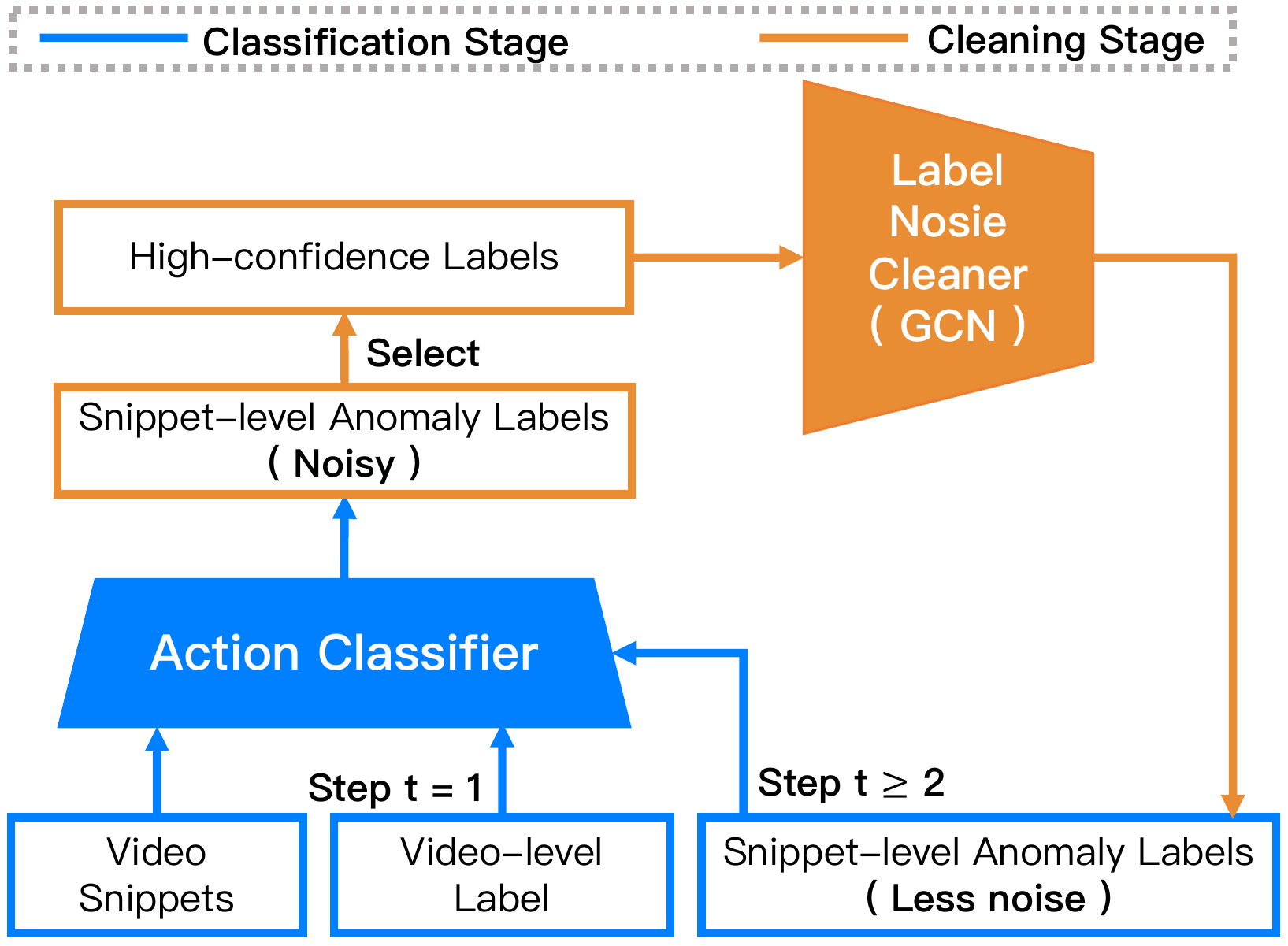}
\caption{\emph{The concept of alternate optimization mechanism.} Noisy labels predicted by the action classifier are utilized to train the label noise cleaner and then they are refined. The cleaned labels are reassigned to optimize the action classifier. The two training processes are executed alternatively.}
\label{fig:EM-Framework}
\end{figure}

Anomaly detection in videos has been long studied for its ubiquitous applications in real-world scenarios, \eg intelligent surveillance, violence alerting, evidence investigation, \etc. Since anomalous events are rarely seen in common environments, anomalies are often defined as behavioral or appearance patterns different from usual patterns in previous work~\cite{benezeth2009co,adam2008monitors,cong2011sparse}. Based on this definition, a popular paradigm for anomaly detection is one-class classification~\cite{xu2015deep,cheng2015cvpr} (a.k.a. unary classification), \ie, to encode the usual pattern with only normal training samples. Then the distinctive encoded patterns are detected as anomalies. However, it is impossible to collect all normal behaviors in a dataset. Therefore some normal events might deviate from the encoded patterns, and could cause false alarms. In recent years, there has been some research~\cite{he2017anomaly,10.1007/978-3-642-32639-4_10,Sultani_2018_CVPR} on an emerging \emph{binary-classification paradigm}: the training data contain both anomalous and normal videos.

Following the binary-classification paradigm, we attempt to address the \emph{weakly supervised anomaly detection} problem, on which only video-level anomaly labels are available in the training data. In this problem, there are neither trimmed anomalous segments nor temporal annotations for the consideration of the human-labor cost. 

The weakly supervised anomaly detection problem is viewed as a multiple-instance learning (MIL) task in prior works~\cite{he2017anomaly,10.1007/978-3-642-32639-4_10,Sultani_2018_CVPR}. They consider a video (or a set of snippets) as a bag, which consists of the snippets (or frames) deemed as instances, and learn instance-level anomaly labels via bag-level annotations. In this paper, we address the problem from a new perspective, formulating it as a supervised learning task under noise labels. The \emph{noise labels} refer to wrong annotations of normal snippets within anomalous videos, since a video labeled as ``anomaly'' may contain quite a few normal clips. In such a viewpoint, we can directly train fully supervised action classifiers once the noisy labels are cleaned. 

There are noticeable advantages of our noise-labeled perspective \emph{in both the training and the test phase}. Instead of simply extracting offline features for MIL models, our action classifier participates in the whole learning process. During the training process, the only difference between action classifiers and fully supervised updating is the input labels. As a result, we preserve all strengths of theses action classifiers, such as well-designed structures, transferable pre-trained weights, ready-to-use source codes, \etc. As for testing, the trained classifier can directly make predictions without any post-processing. It is extremely convenient and highly efficient because the feature extraction and the abnormality decision are seamlessly integrated into a single model.

Intuitively, a well-trained classifier yields the predictions with less noise, and the cleaned labels in turn help to train a better classifier. To this end, we design an alternate training procedure as Figure~\ref{fig:EM-Framework} illustrates. It consists of two alternate stages, \ie, cleaning and classification. In the cleaning stage, we train a cleaner to correct the noisy predictions obtained from the classifier, and the cleaner provides refined labels with less noise. In the classification stage, the action classifier is retrained with the cleaned labels and generates more reliable predictions. Such a cyclic operation is executed several times until convergence. The main idea of our cleaner is to eliminate noise of low-confidence predictions via high-confidence ones. We devise a graph convolutional network (GCN) to establish relationships between high-confidence snippets and low-confidence ones. In the graph, snippets are abstracted into vertexes and the anomaly information is propagated through edges. During testing, we no longer require the cleaner and directly obtain snippet-wise anomaly results from the trained classifier. For verification of the general applicability of our model, we carry out extensive experiments with two types of mainstream action classifiers: a 3D-conv network C3D~\cite{tran2015learning} and a two-stream structure TSN~\cite{tsn_wang}. In addition, we evaluate the proposed approach on 3 different-scale datasets, \ie, {UCF-Crime}~\cite{Sultani_2018_CVPR}, {ShanghaiTech}~\cite{Luo_2017_ICCV} and {UCSD-Peds}~\cite{li2014crowded}. The experimental results demonstrate that our model advances the state-of-the-art performance of weakly supervised anomaly detection. 

In a nutshell, the contribution of this paper is three-fold:
\begin{itemize}
  \item We formulate the problem of anomaly detection with weak labels as a supervised learning task under noise annotations, and put forward an alternate training framework to optimize the action classifier.
  \item	We propose a GCN to clean noise labels. To the best of our knowledge, it is the first work to apply a GCN to correct label noise in the area of video analytics.
  \item	We conduct experiments on 3 different-scale anomaly detection datasets with two types of action classifiers, in which the state-of-the-art performance validates the effectiveness of our approach. The source code is available at {\color{blue}\href{https://github.com/jx-zhong-for-academic-purpose/GCN-Anomaly-Detection}{https://github.com/jx-zhong-for-academic-purpose/GCN-Anomaly-Detection}}.
\end{itemize}

%-------------------------------------------------------------------------
\section{Related Work}
\textbf{Anomaly detection.} As one of the most challenging problem, anomaly detection in videos has been extensively studied for many years~\cite{kratz2009cvpr,zhao2011sparse,wu2010chaotic,hasan2016regularity,li2014multiscale,anti2011parse,mohammadi2016angry,li2014crowded} . Most research addresses the problem under the assumption that anomalies are rare or unseen, and behaviors deviating from normal patterns are supposed to be anomalous. They attempt to encode regular patterns via a variety of statistic models, \eg the social force model~\cite{mehran2009social}, the mixture of dynamic models on texture~\cite{li2014crowded}, Hidden Markov Models on video volumes~\cite{hospedales2009hmm,kratz2009cvpr}, the Markov Random Field upon spatial-temporal domain~\cite{kim2009mrf}, Gaussian process modeling~\cite{li2015gaussian,cheng2015cvpr}, and identify anomalies as outliers. 
%Adam \etal~\cite{adam2008monitors} set a grid of monitoring site on the image, and detect anomalies at each site via comparing a histogram calculated from optical flow. 
Sparse reconstruction~\cite{lu2013matlab,luo2017sparse,cong2011sparse,zhao2011sparse} is also another popular approach for usual pattern modeling.
%Cong \etal~\cite{cong2011sparse} 
They utilize sparse representation to construct a dictionary for normal behavior, and detect anomalies as the ones with high reconstruction error. Recently, with the great success of deep learning, a few researchers design deep neural networks on abstraction feature learning~\cite{hasan2016regularity,chong2017autoencoder,luo2017history} or video prediction learning~\cite{liu2017future} for anomaly detection. %Hasan \etal~\cite{hasan2016regularity}utilize a 3D convolutional Auto-Encoder network for regular pattern modeling. 
As opposed to the works that built their detection models on normal behavior only, there is research~\cite{adhiya2009tracking,he2017anomaly,Sultani_2018_CVPR} employing both usual and unusual data for model building. Among them, MIL is used for motion pattern modeling under weakly supervised setting~\cite{he2017anomaly,Sultani_2018_CVPR}. Sultani \etal~\cite{Sultani_2018_CVPR} propose an MIL-based classifier to detect anomalies, where a deep anomaly ranking model predicts anomaly scores. Unlike them, we formulate the anomaly detection problem with weak labels as a supervised learning under noise labels, and devise an alternate training procedure to progressively promote the discrimination of action classifiers.

\textbf{Action analysis.} Action classification is a long standing problem in the field of computer vision, and a large body of research works~\cite{wang2013action,tran2015learning,tsn_wang,carreira2017quo,karpathy2014large,Wang_2018_ECCV} have been presented. A majority of modern approaches have introduced deep architecture models~\cite{carreira2017quo,tran2015learning,simonyan2014two,tsn_wang}, including the most prevailing two-stream networks ~\cite{simonyan2014two}, C3D~\cite{tran2015learning} and their variants~\cite{tsn_wang,feichtenhofer2016convolutional,qiu2017learning,carreira2017quo}. Up to now, deep learning based methods have achieved state-of-the-art performance. Besides action classification, some researchers recently have focused on temporal action localization~\cite{zhao2017ssn,Lin:2017:SST:3123266.3123343,zjx_mm2018,zheng_eccv18_autoloc,gao2017cascaded}. The performance metrics of temporal action detection and anomaly detection are quite different: action detection aims to find a temporal interval overlapped with the ground truth as much as possible, whereas anomaly detection aims for a robust frame-level performance under various discrimination thresholds. In this paper, we attempt to leverage the powerful action classifiers to detect anomalies in a simple and feasible way.

\textbf{Learning under noisy labels.} The research works~\cite{larsen1998design,natarajan2013learning,patrini2017making,goldberger2017training} addressing the noise label problem can be generally divided into two categories: noise reduction and loss correction. In the case of noise reduction, they aim to correct noisy labels via formulating the noise model explicitly or implicitly, such as Conditional Random Fields (CRF)~\cite{vahdat2017toward}, knowledge graphs~\cite{li2017learning}. Approaches in the latter group are developed for directly learning with label noise, utilizing correction methods for loss adjustment. Azadi \etal~\cite{azadi2016auxiliary} actively select training features via imposing a regularization term on loss function. Different from theses general approaches, our GCN is intended for videos and take advantages of the video-based characteristics.

\textbf{Graph convolutional neural network.} In recent years, a surge of graph convolutional networks~\cite{ou2016asymmetric,Kipf2016Semi,Pfeiffer:2014:AGM:2566486.2567993,li2015gated,grover2016node} have been proposed to tackle graph-structured data. An important stream of these works is utilizing spectral graph theory~\cite{bruna2014spetral,defferrard2016convolutional}, which decomposes the graph signal on the spectral domain and defines a series of parameterized filters for convolution. A number of researchers propose improvements of spectral convolutions, leading to advanced performances on tasks such as node classification and recommendation system. The goal of our label noise cleaner is classifying nodes (video snippets) in a graph (the whole video) under the supervision of high-confidence annotations.

%------------------------------------------------------------------------
\section{Problem Statement}\label{sec:problem_statement}
Given a video \(V=\{v_i\}_{i=1}^{N}\) with \(N\) snippets, the observable label \(Y \in \{1,0\}\) indicates whether this video contains anomalous clips or not. Note that no temporal annotation is provided in training data. The goal of anomaly detection is to pinpoint the temporal position of abnormalities once they occurs in test videos.

Sabato and Tishby~\cite{Sabato:2012:MLA:2503308.2503338} provide a theoretical analysis in which MIL tasks can be viewed as learning under one-sided label noise. In some prior works~\cite{he2017anomaly,10.1007/978-3-642-32639-4_10,Sultani_2018_CVPR}, anomaly detection under the weak supervisory signal is described as a typical MIL problem. Therefore, we naturally cast anomaly detection from MIL formulation to noisy label setting.

\textbf{MIL formulation.}
In this formulation, each clip \(v_i\) is considered as an \textit{instance}, of which the anomaly label \(y_i\) is unavailable. These clips compose the \textit{positive/negative bag} according to the given video-level anomaly label \(Y\): a \textit{positive bag} \((Y=1)\) includes at least one anomalous clip, while a \textit{negative bag} \((Y=0)\) is entirely comprised of normal snippets. Consequently, anomaly detection is modeled as key instance detection~\cite{liu2012key} under MIL, in search of positive instances \(v_i\) with \(y_i=1\). This MIL setting allows learning instance-level labels under bag-level supervision, and a set of approaches~\cite{he2017anomaly,10.1007/978-3-642-32639-4_10,Sultani_2018_CVPR} is derived from this.

\textbf{Noisy-labeled learning formulation.}
It is evident that the label \(Y=0\) is noiseless, since it means all snippets \(v_i\) in the video \(V\) are normal:
\begin{equation}
    Y=0 \Rightarrow y_i=0,\ \forall v_i \in V \,.
\end{equation}
However, \(Y=1\) is noisy because in this case the video \(V\) is partially made up of anomalous clips:
\begin{equation}
   Y=1 \not\Rightarrow y_i=1,\ \forall v_i \in V \,.
\end{equation}
This is referred to as \textit{one-sided label noise}~\cite{Blum1998,CARBONNEAU2018329,pmlr-v30-Scott13}, for the noise only appears along with \(Y=1\). As long as appropriately handling the label noise w.r.t. \(Y=1\), we are able to readily apply a variety of well-developed action classifiers to anomaly detection.

\section{Graph Convolutional Label Noise Cleaner}
\begin{figure*}[!t]
\centering\includegraphics[width=1.0\textwidth]{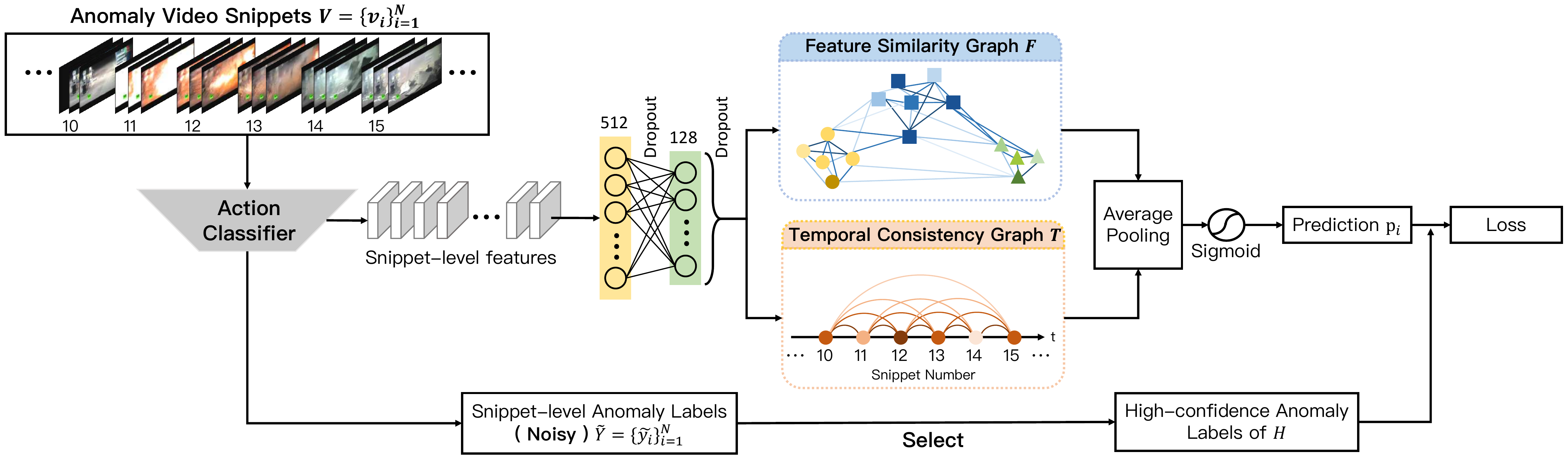}
\caption{\textit{Overview of the training process of label noise cleaner.} The action classifier extracts spatio-temporal features from anomalous video snippets and outputs noisy snippet-level labels. Snippet-level features from the classifier are compressed and fed into two graph modules to model the feature similarity and temporal consistency of snippets. In the two graph-based modules, A darker node represents higher anomaly confidence of the snippet. The output of these two models are fused and utilized to predict the snippet-level labels with less noise. The loss is updated to correct the predictive noise via high-confidence snippets.}
\label{fig:cleaner}
\end{figure*}
% We train the two graph models utilizing

Similar to many noisy-labeled learning approaches, our method adopts an EM-like optimization mechanism: alternately training the action classifier and the noise cleaner. At each training step of the noise cleaner, we have obtained rough snippet-wise anomaly probabilities from the action classifier, and the target of our noise cleaner is to \emph{correct low-confidence anomaly scores via high-confidence ones.} 

Unlike other general noise-labeled learning algorithms, our cleaner is specifically designed for videos. To the best of our knowledge, this is the first work to deploy a GCN in noise-labeled videos. In the graph convolutional network, we leverage two characteristics of a video to correct the label noise, \ie, \textit{feature similarity} and \textit{temporal consistency}. Intuitively, \textit{feature similarity} means the anomaly snippets share some similar characteristics, while \textit{temporal consistency} means anomaly snippets probably appear in temporal proximity of each other.

\subsection{Feature Similarity Graph Module}
As Figure~\ref{fig:cleaner} depicts, features from the action classifier are first compressed with two fully connected layers to mitigate the curse of dimensionality~\cite{Bellman:1957}. We model the feature similarly with an attributed graph~\cite{Pfeiffer:2014:AGM:2566486.2567993} \(F=(V, E, \bf{X})\) , where \(V\) is the vertex set, \(E\) is the edge set, and \(\bf X\) is the attribute of vertexes. In particular, \(V\) is a video as defined in Section~\ref{sec:problem_statement}, \(E\) describes the feature similarity amongst snippets, and \(\bf{X} \in \mathbb{R}^{N \times d}\) represents the \(d\)-dimensional feature of these \(N\) snippets. The adjacency matrix \({\bf{A^F}} \in \mathbb{R}^{N \times N}\) of \(F\) is defined as:
\begin{equation}
   %{\bf{A^F}}_{(i,j)}=1+cos({\bf{X}}_i, {\bf{X}}_j) \,,
   {\bf{A^F}}_{(i,j)}=\exp({\bf{X}}_i \cdot {\bf{X}}_j - max({\bf{X}}_i \cdot {\bf{X}})) \,,
\end{equation}
where the element \({\bf{A^F}}_{(i,j)}\) measures the feature similarly between the \(i^{th}\) and \(j^{th}\) snippets. Since an adjacency matrix should be non-negative, we bound the similarity to the range \((0,1]\) with a normalized exponential function. Based on the graph \(F\), snippets with similar features are closely connected, and the label assignments are propagated differently in accordance with different adjacency values.

The nearby vertexes are driven to have the same anomaly label via graph-Laplacian operations. Following Kipf and Welling~\cite{Kipf2016Semi}, we approximate the graph-Laplacian with a renormalization trick:
\begin{equation}\label{adj_renorm}
   {\bf\widehat{A}^F}={\widetilde{\bf D}^{{\bf F}-\frac{1}{2}}} {\bf\widetilde{A}^F} {\widetilde{\bf D}^{{\bf F}-\frac{1}{2}}} \,,
\end{equation}
where the self-loop adjacency matrix \({\bf\widetilde{A}^F}={\bf{A^F}}+{\bf I_n}\), and \({\bf I_n} \in \mathbb{R}^{N \times N}\) is the identity matrix; \({\bf\widetilde{D}^F}_{(i,i)}=\sum_j {{\bf\widetilde{A}^F}_{(i,j)}}\) is the corresponding degree matrix. Finally, the output of a feature similarity graph module layer is computed as:
\begin{equation}
   {\bf H^F}= \sigma({\bf\widehat{A}^F} {\bf X} {\bf W}) \,,
\end{equation}
where \(\bf W\) is a trainable parametric matrix, and \(\sigma\) is an activation function. Since the whole computational procedure is differentiable, our feature similarity graph module can be trained in an \textit{end-to-end} fashion. Therefore, neural networks are capable of seamlessly incorporating the single or multiple stacked modules. Although the aforementioned procedure contains some element-wise calculations, we provide a high-efficient vectorized implementation in \textbf{Appendix}.

Recently, Wang and Gupta~\cite{Wang_2018_ECCV} also have established similarity graphs to analyze a video. Nevertheless, both the goal and the method are quite different from ours: they aim to capture long-term dependencies with the similarity relations of correlated objects/regions, whereas we attempt to propagate supervisory signals with the similarity levels of entire snippets/frames.

\subsection{Temporal Consistency Graph Module}
As pointed out in~\cite{jayaraman2016slow,mobahi2009deep,wiskott2002slow}, temporal consistency is advantageous to many video-based tasks. The temporal consistency graph \(T\) is directly built upon the temporal structure of a video. Its adjacency matrix \({\bf{A^T}} \in \mathbb{R}^{N \times N}\) is only dependent on temporal positions of the \(i^{th}\) and \(j^{th}\) snippets:
\begin{equation}\label{eq:adj_temp}
   {\bf{A^T}}_{(i,j)}=k(i, j) \,,
\end{equation}
where \(k\) is a non-negative kernel function. Consider that the kernel is supposed to distinguish various temporal distances and closely connect the snippets in vicinity. In practice, we use an exponential kernel (a.k.a. Laplacian kernel) neatly bounded in \((0,1]\):
\begin{equation}
   k(i, j)=\exp(-||i-j||) \,.
\end{equation}

Likewise, we obtain the renormalized adjacency matrix \(\bf\widehat{A}^T\) as {Equation~\ref{adj_renorm}} for the graph-Laplacian approximation, and the forward result of this module is computed as:
\begin{equation}
   {\bf H^T}= \sigma({\bf\widehat{A}^T} {\bf X} {\bf W}) \,,
\end{equation}
where \(\bf W\) is a trainable parametric matrix, \(\sigma\) is an activation function, and \(\bf X\) is the input feature matrix. The stacked temporal consistency graph layers also can be conveniently included into neural networks.

\subsection{Loss Function} 

Finally, the outputs of the above two modules are fused with an average pooling layer, and activated by a Sigmoid function to make the probabilistic prediction \(p_i\) of each vertex in the graph, corresponding to the anomaly probability of our noise cleaner w.r.t. the \(i^{th}\) snippet. The loss function \(\mathcal{L}\) is based upon two types of supervision:
\begin{equation}
   \mathcal{L} = \mathcal{L}_D + \mathcal{L}_I \,,
\end{equation}
where \(\mathcal{L}_D\) and \(\mathcal{L}_I\) are computed under the \textit{direct} and the \textit{indirect} supervision respectively. Given the rough snippet-wise anomaly probabilities \(\widetilde{Y}=\{\widetilde{y}_i\}_{i=1}^{N}\) from the action classifier. The loss term under \textit{direct supervision} is defined as a cross-entropy error over the high-confidence snippets:
\begin{equation}\label{eq:H}
   \mathcal{L}_D = -\frac{1}{|H|}\sum_{i \in H} [\widetilde{y}_i\ln{p_i}+(1-\widetilde{y}_i)\ln{(1-p_i)}] \,,
\end{equation}
where \(H\) is the set of high-confidence snippets. We oversample each video frame with the ``10-crop'' augment,\footnote{``10-crop'' means cropping images into the center, four corners, and their mirrored counterparts.} and calculate mean anomaly probabilities \(\widetilde{y}_i\) as well as predictive variances of the action classifier. As pointed out by Kendall and Gal~\cite{NIPS2017_7141}, variance measures the uncertainty of predictions. In other words, \textit{the smaller variance indicates the higher confidence.} This criterion of confidence is conceptually simple yet practically effective.

The \textit{indirectly supervised term} is a temporal-ensembling strategy~\cite{temp_ensemble} to further harness a small number of labeled data, because high-confidence predictions are only from a portion of the entire video. Its main idea is to smooth the network predictions
of all snippets at different training steps:
\begin{equation}
   \mathcal{L}_I = \frac{1}{N} \sum_{i=1}^N |p_i-\overline{p}_i| \,,
\end{equation}
where \(\overline{p}_i\) is the discount-weighted average predictions of our noise cleaner over various training epochs. There is a major difference between the original ``cool start'' initialization and our implementation as explained in \textbf{Appendix}, since we have already obtained a set of rough predictions from the action classifier.

\subsection{Alternate Optimization}
The training process of our noise cleaner is merely one part of the alternate optimization. The other part, \ie, the training process of our classifier, is exactly the same as common fully supervised updating, \emph{except that the labels are snippet-wise predictions from our trained cleaner}. After repeating such an alternate optimization several times, final anomaly detection results are directly predicted by the last trained classifier. Obviously, almost no change in the action classifier is required during the training or the test phase. As a result, we can conveniently train the fully supervised action classifier under weak labels, and directly deploy it for anomaly detection without all the bells and whistles.

\section{Experiments}\label{sec_experiment}

\subsection{Datasets and Evaluation Metric}
We conduct the experiments upon three datasets of various scales, \ie, {UCF-Crime}~\cite{Sultani_2018_CVPR}, {ShanghaiTech}~\cite{Luo_2017_ICCV} and {UCSD-Peds}~\cite{li2014crowded}. 

\textbf{{UCF-Crime}} is a large-scale dataset of real-world surveillance videos. It has 13 types of anomalies with 1,900 long untrimmed videos, which consist of 1,610 training videos and 290 test videos.

\textbf{{ShanghaiTech}} is a medium-scale dataset of 437 videos, including 130 abnormal events on 13 scenes. In the standard protocol~\cite{Luo_2017_ICCV}, all training videos are normal, and this setting is inappropriate for the binary-classification task. Hence, we reorganize the dataset by randomly selecting anomaly testing videos into training data and vice versa. Meanwhile, both training videos and testing ones cover all of the 13 scenes. This new split of the dataset will be available for follow-up comparisons. More details are given in \textbf{Appendix}.

\textbf{{UCSD-Peds}} is a small-scale dataset made up of two subsets: Peds1 has 70 videos, and Peds2 has 28 videos. Since the former is more frequently used for pixel-wise anomaly detection~\cite{xu2015deep}, we only conduct experiments on the latter as in~\cite{Luo_2017_ICCV}. Similarly, the default training set does not contain anomaly videos. Following He \etal~\cite{he2017anomaly}, 6 anomaly videos and 4 normal ones on UCSD-Peds2 are randomly included into training data, and the remaining videos constitute the test set. We also repeat this process 10 times and report the average performance.

\textbf{Evaluation Metric.} Following previous works~\cite{Luo_2017_ICCV,he2017anomaly,Sultani_2018_CVPR}, we plot the frame-level receiver operating characteristics (ROC) curve and compute an area under the curve (AUC) as the evaluation metric. In the task of temporal anomaly detection, a larger frame-level AUC implies the higher diagnostic ability, as well as the robuster performance at various discrimination thresholds.

\begin{table*}[!h]\small
\begin{floatrow}
\capbtabbox{
 \begin{tabular}{ccccccc}
    \hline
    \bf Training &\bf Indirect & \multicolumn{2}{c}{\bf Temporal}    & \multicolumn{2}{c}{\bf Feature}    & {\bf AUC} \\ %\multirow{3}{*}{\bf AUC} \\
  \bf Stage &\bf Supervision  & \multicolumn{2}{c}{\bf Consistency} & \multicolumn{2}{c}{\bf Similarity} &     {\bf (\%)}\\ \cline{3-6}
   
       & & Conv.           & Graph         & Conv.          & Graph         &     \\
        \hline\hline
      Step-2  &   $\surd$      &       $\surd$        &   $\surd$             &  $\surd$             &  $\surd$ &  74.60\\
      \hline
Step-2   &         &       $\surd$        &   $\surd$             &  $\surd$             &  $\surd$ &  73.79\\
Step-2   &         &       $\surd$        &                &               &   &  67.57\\
Step-2   &         &       $\surd$        &    $\surd$            &               &   &  72.93\\
Step-2   &         &               &                &  $\surd$             &   &  67.23\\
Step-2   &         &               &                &  $\surd$             &  $\surd$ &  72.44\\
      \hline
      Step-1  &     --        &   --         &    --        &      --     &  -- &  70.87 \\
    \hline
   \end{tabular}
}{
 \caption{\emph{Ablation Studies on {UCF-Crime}.}}
 \label{tab:ucf_ablation}
}
\hspace{-2em}
\capbtabbox{
  \begin{tabular}{ccc}
    \hline
    \textbf{Method} & \textbf{AUC (\%)} & \textbf{False Alarm (\%)} \\ %
    \hline\hline
    SVM Baseline & 50.0 & -- \\
    Hasan \etal\cite{hasan2016regularity}  & 50.6 & 27.2 \\
    Lu \etal\cite{lu2013matlab}  &  65.51 & 3.1 \\
    Sultani \etal$^\dagger$~\cite{Sultani_2018_CVPR}  &  74.44 & -- \\
    Sultani \etal$^\ddagger$~\cite{Sultani_2018_CVPR}  & 75.41 & 1.9 \\
    \hline
    \textbf{Ours}\\
    C3D                & 81.08 & 2.8 \\
    TSN$^{RGB}$          & 82.12 & 0.1 \\
    TSN$^{Optical Flow}$ & 78.08 & 1.1 \\
    \hline
  \end{tabular}
}{
 \caption{\emph{Quantitative comparison on {UCF-Crime}.} $\dagger$ and $\ddagger$ indicate the loss without and with constraints respectively. } 
 \label{tab:ucf_det}
}
\end{floatrow}
\end{table*}

\subsection{Implementation Details}

\textbf{Action classifiers.} For verification of the general applicability of our model, we utilize two mainstream structures of action classifiers in the experiments. \textbf{C3D}~\cite{tran2015learning} is a \emph{3D-convolutional} network. The model is pre-trained on the Sports-1M~\cite{karpathy2014large} dataset. In the training process, we input features from its \textit{fc7} layer into our label noise cleaner. \textbf{Temporal Segment Network (TSN)}~\cite{tsn_wang} is a \emph{two-stream} architecture. We choose BN-Inception~\cite{ioffe2015batch} pre-trained on Kinetics-400~\cite{carreira2017quo} as the backbone, and extract features from its \textit{global\_pool} layer to train our noise cleaner. The action classifiers are both implemented upon the Caffe~\cite{Jia:2014:CCA:2647868.2654889} platform with the same settings of video sampling and data augment as~\cite{tsn_wang}. In all the experiments, we keep the default settings if not specified particularly.

\textbf{Label noise cleaner.} After we add the author list and the acknowledgement section into our camera-ready version, this part has to be moved to \textbf{Appendix} because of limited space. Please refer to our Github page and \textbf{Appendix}.

\begin{figure}[h]
\includegraphics[width=0.75\textwidth]{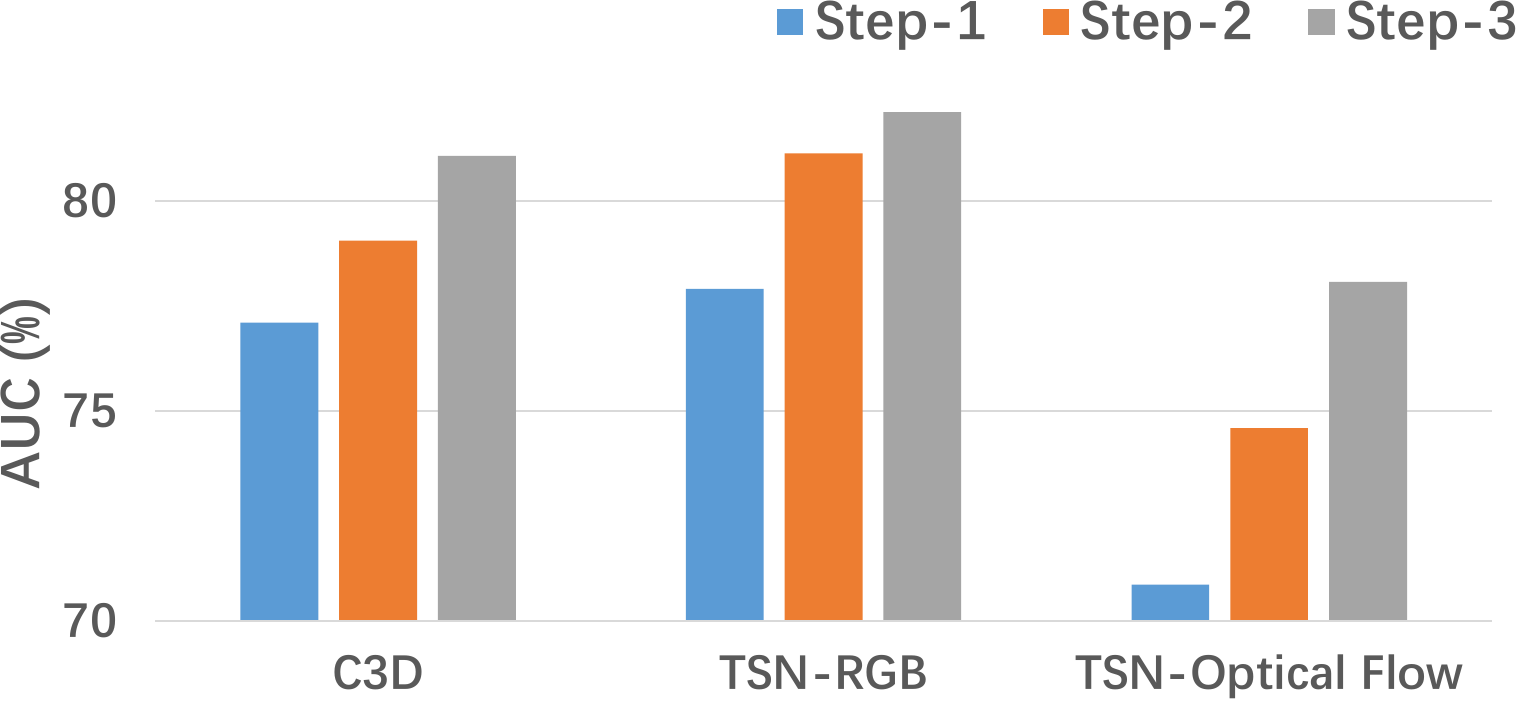}
\caption{\emph{Step-wise performance on {UCF-Crime}.}}
\label{fig:ucf_step}
\end{figure}

\subsection{Experiments on UCF-Crime}

Under the video-level supervision, we train C3D with 18,000 iterations. As for TSN, the initial iteration number of both streams is 20,000. At each re-training step, we stop the updating procedure at 4,000 iterations.

\textbf{Step-wise results.}
As Figure~\ref{fig:ucf_step} depicts, we report the AUC performance at each step to evaluate the efficacy of our alternate training mechanism. Even if only given video-level labels, C3D and the RGB branch of TSN can achieve a descent performance at the Step-1. It is a wise choice for us to involve action classifiers in the training process. However, the optical flow stream of TSN is far from satisfaction, which reflects the necessity of our noise cleaner. At the following steps, the proposed approach significantly improves the detection performance of all the action classifiers. Faced with the most noise in initial predictions, the AUC performance of our optical flow branch is still boosted from 70.87\% to 78.08\% with a relative gain of 10.2\%.

\textbf{Indirect supervision.}
We conduct ablation studies upon the optical flow modality of TSN. First, we exclude the indirectly supervised term from the loss to verify its effectiveness. As on the \(2^{nd}\) row of Table~\ref{tab:ucf_ablation}, the performance slightly declines from 74.60\% to 73.79\%, but the gain on the result of Step-1 remains considerable. In the following ablations, we remove the indirect supervised term to eliminate interference.

\textbf{Temporal consistency.}
We would like to explore two questions: \emph{Is temporal information helpful? Can our graph convolution utilize this information?} By excluding the other interference factors, there is only the temporal consistency module. To remove the graph of temporal information, we fill the \(\bf A^T\) in Equation~\ref{eq:adj_temp} with 0.5 (the mid-value of its bounds) and reproduce the alternate training procedure. As shown on the \(3^{rd}\) row of Table~\ref{tab:ucf_ablation}, the performance without temporal graph is worse than that of Step-1, in which case the GCN only memorizes the pattern of high-confidence predictions but ignores other snippets. As for the ablation on graph convolution, we observe that the independent temporal consistency module boosts the AUC to 72.93\% as on the \(4^{th}\) row of Table~\ref{tab:ucf_ablation}, which demonstrates that our graph convolution really capitalizes on the temporal information.

\begin{figure}[!ht]
\centering
\includegraphics[width=0.7\textwidth]{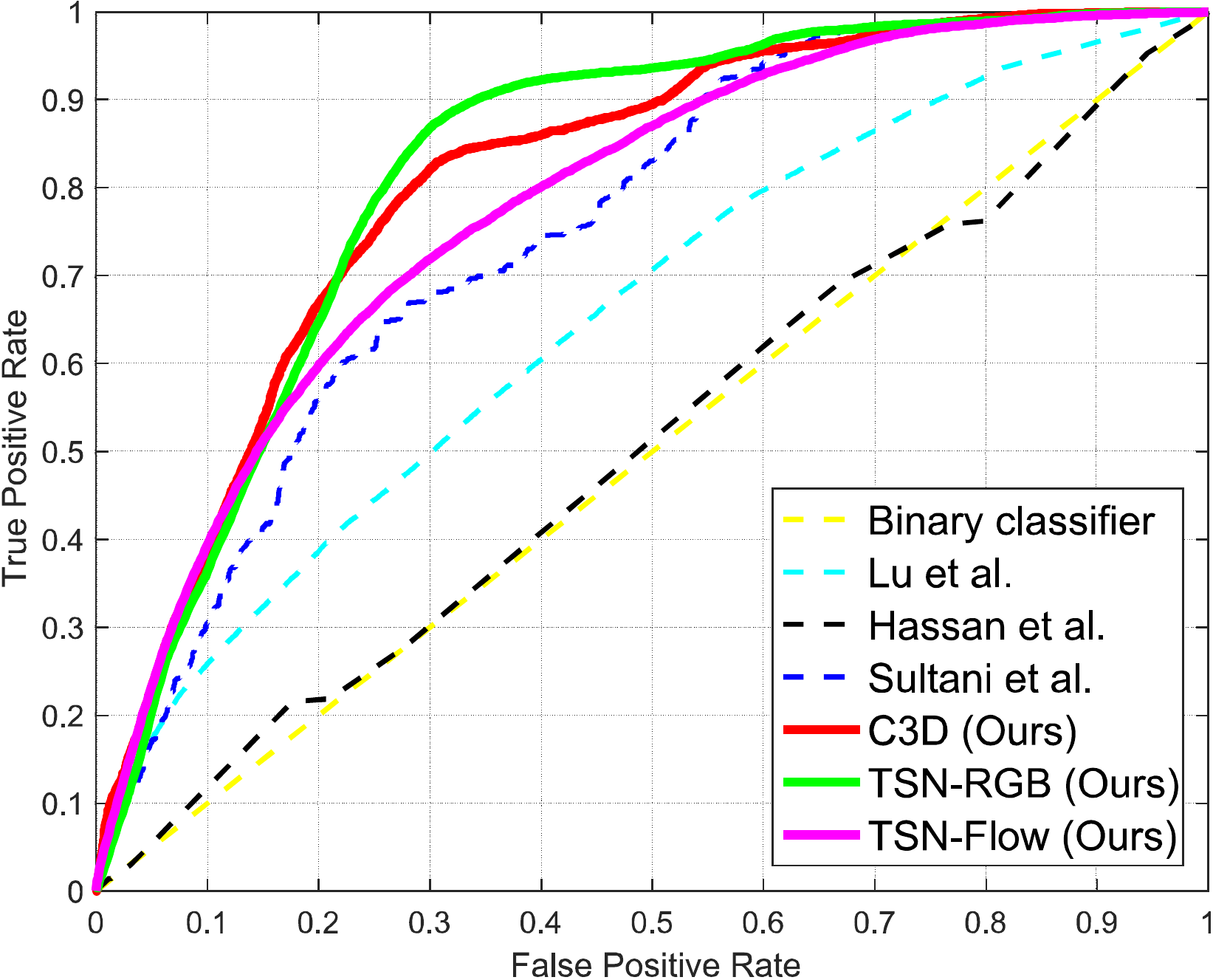}
\caption{\emph{ROC curves on UCF-Crime}.}
\label{fig:ucf_roc}
\end{figure}

\textbf{Feature similarity.} Likewise, we only reserve the feature similarity module to investigate the efficacy of similarity graphs and our convolutional operation. We first damage the feature similarity graph by setting all elements of the adjacency matrix as the mid-value. As on the \(5^{th}\) row of Table~\ref{tab:ucf_ablation}, the AUC value falls to 67.23\% without the graph. After recovering the original feature similarity graph, the single feature similarity module can increase the AUC value from 70.87\% to 72.44\% as shown on the \(6^{th}\) row of Table~\ref{tab:ucf_ablation}. This illustrates that both similarity graphs and the convolution are beneficial to clean the noisy labels.

%\textbf{When should we stop?}

\begin{figure*}[h]
  \centering
  
  \subfloat[Burglary005]{
  	\includegraphics[width=0.23\textwidth]{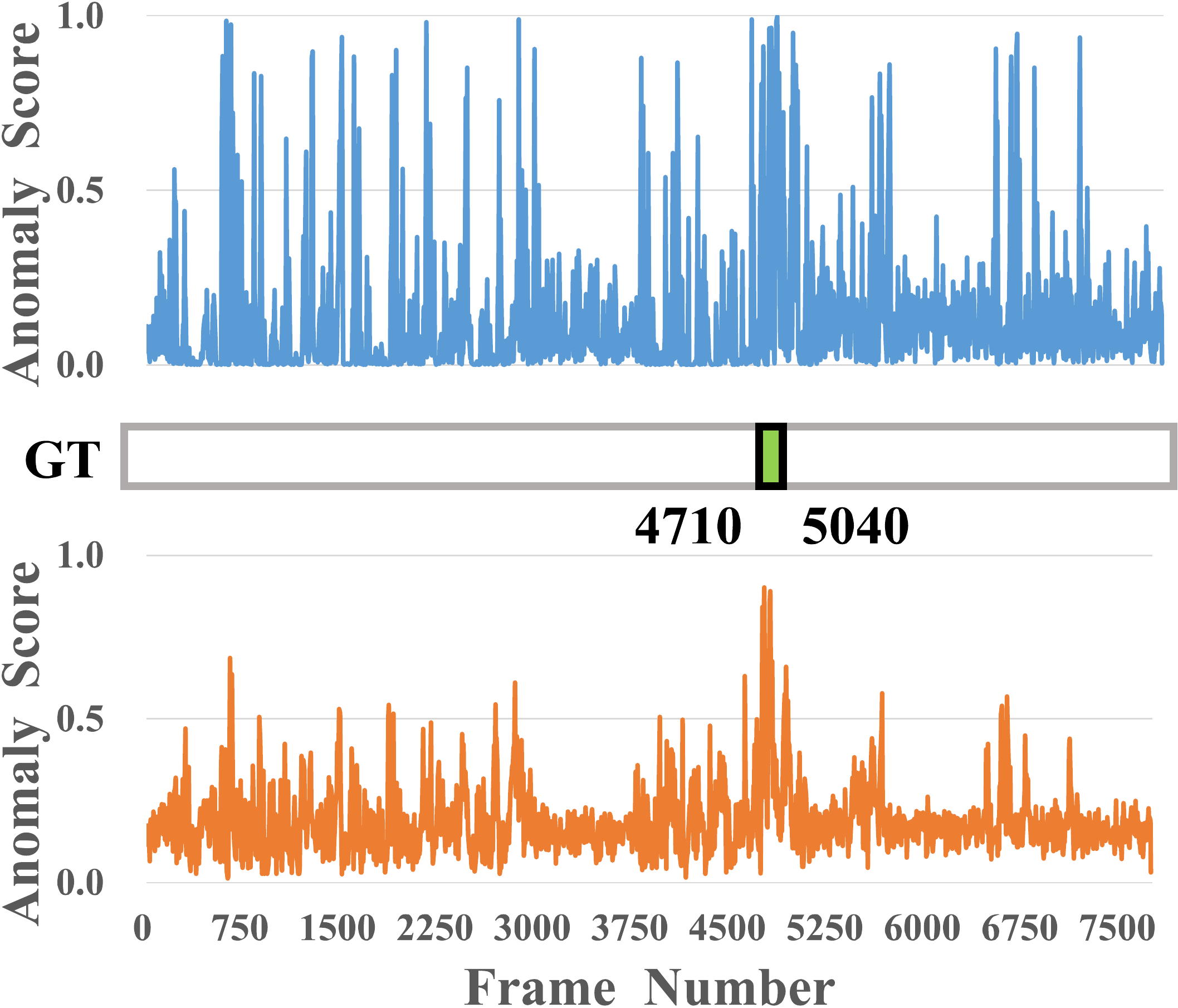}
  }
  \subfloat[Burglary079]{
    \includegraphics[width=0.23\textwidth]{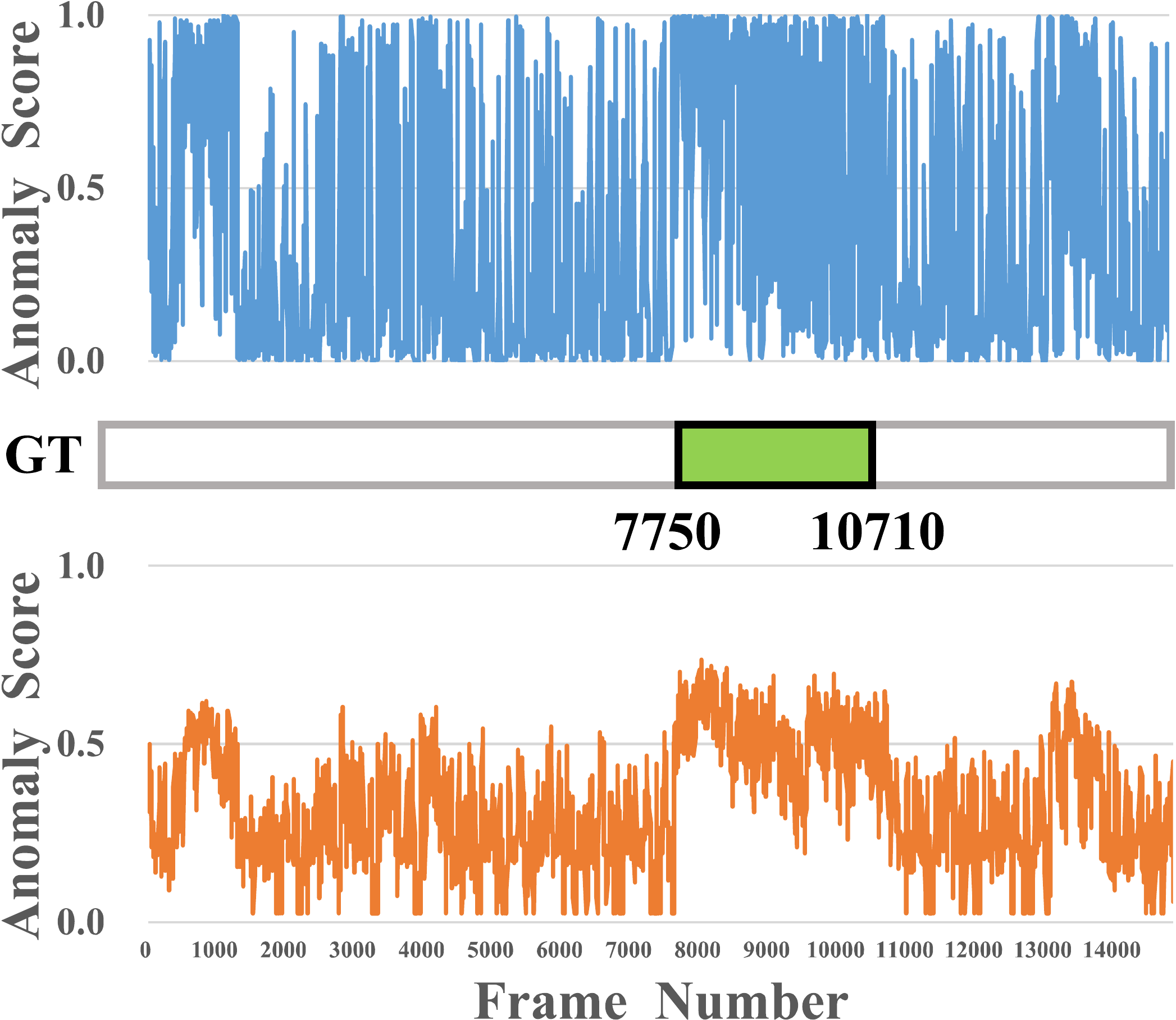}
    }
  \subfloat[Arrest007]{
    \label{fig:subfig:Arrest007_result} 
    \includegraphics[width=0.23\textwidth]{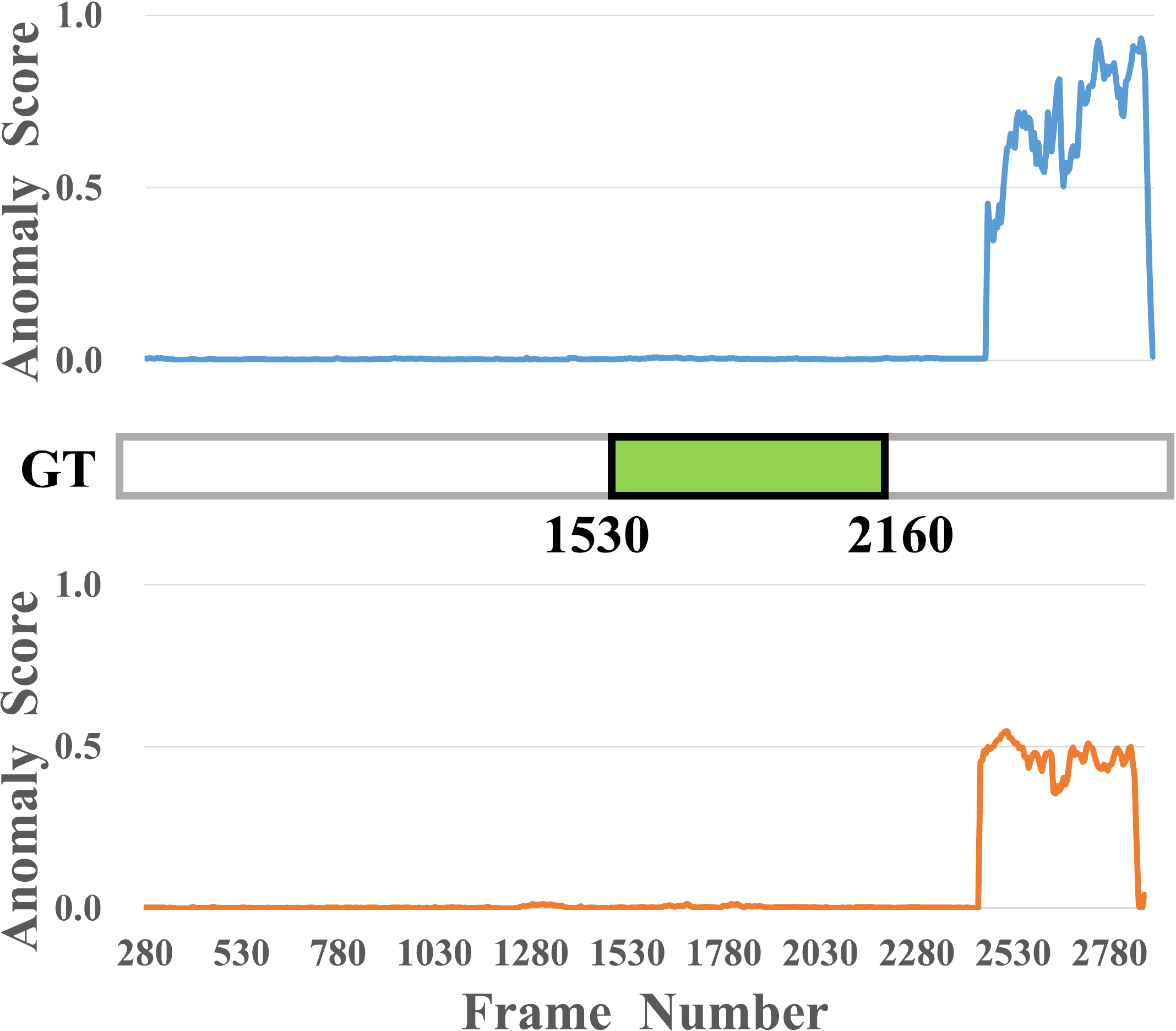}
    }
  \subfloat[Arrest007 (Partial Ground Truth)]{
  	\animategraphics[loop,autoplay,poster=17,width=0.23\textwidth]{12}{figures/Arrest007RGB/image_00}{1500}{1600}
  	\label{fig:subfig:Arrest007_video} 
   }
    
%   \vspace{-1.0em}
  \caption{\emph{Visualization of testing results on UCF-Crime.} The blue curves are predictions of the action classifier trained under video-level labels, and the orange curves are the results under cleaned supervision. The ``GT'' bars in green are ground truths. {\color{cyan}{Best viewed in Adobe Reader where (d) should play as a video.}}} 
  \label{fig:ucf_qual} 
\end{figure*}

\textbf{Quantitative comparison.}
We compare our methods with state-of-the-art models upon 3 indicators, \ie, ROC curves, AUC and false alarm rates. As Figure~\ref{fig:ucf_roc} shows, our curves of all the action classifiers almost completely enclose the others, which means they are consistently superior to their competitors at various thresholds. The smoothness of the three curves shows the high stability of our proposed approach. As shown in Table~\ref{tab:ucf_det}, we boost the AUC value up to 82.12\% at most. As for false alarm rates at 0.5 detection score, the C3D is slightly inferior to Sultani \etal, whereas the other two classifiers are fairly satisfactory as shown in Table~\ref{tab:ucf_det}. Notably, the RGB branch of TSN reduces the false alarm rate to 0.1\%, nearly 1/20 of the best-so-far result.  

\textbf{Qualitative analysis on the test set.}
To observe the influence of our model, we visualize the before-and-after change in predictions of action classifiers. As presented in Figure~\ref{fig:ucf_qual}, our denoising process substantially alleviates the predictive noise of action classifiers within both normal and anomaly snippets. Intriguingly, the classifier fails to detect the anomaly event in the ``Arrest007'' video from beginning to end as Figure~\ref{fig:subfig:Arrest007_result} depicts. After watching all videos of the ``Arrest'' class, we finally discover the possible cause: the similar scene in this testing video does not exist in training data. In this video, a man is arrested at the laundromat for vandalism of washing machines as shown in Figure~\ref{fig:subfig:Arrest007_video}, while ``Arrest'' events occur on the highway or at the checkout counter in training data. It implies that to detect anomalous events in generic scenes is still a big challenge for the limited generalization ability of current models.  

\subsection{Experiments on ShanghaiTech}

\begin{table}\centering
  \caption{\emph{Step-wise AUC (\%) on ShanghaiTech}.}\label{tab:SH_det}
  \begin{tabular}{cccc}
    \hline
    \textbf{Action Classifier}  &  \textbf{C3D} & \textbf{TSN}$^\textbf{{RGB}}$ & \textbf{TSN}$^\textbf{{Optical Flow}}$\\
    \hline
    {Step-1}       & 73.79 & 80.83 & 78.23 \\
    {Step-2}       & 76.16 & 82.17 & 84.19 \\
    {Step-3}       & 76.44 & 84.44 & 84.13 \\
    \hline
  \end{tabular}
\end{table}

\textbf{Step-wise results.}
As illustrated in Table~\ref{tab:SH_det}, the performance is improved after the alternate training w.r.t. all the action classifiers. The results of optical flow branch of TSN at Step-3 reflects that excessive iterations may deteriorates on the detection performance. Nevertheless, our method performs robustly as the AUC value only drops slightly.

\begin{figure}[ht]
\centering
\includegraphics[width=0.9\textwidth]{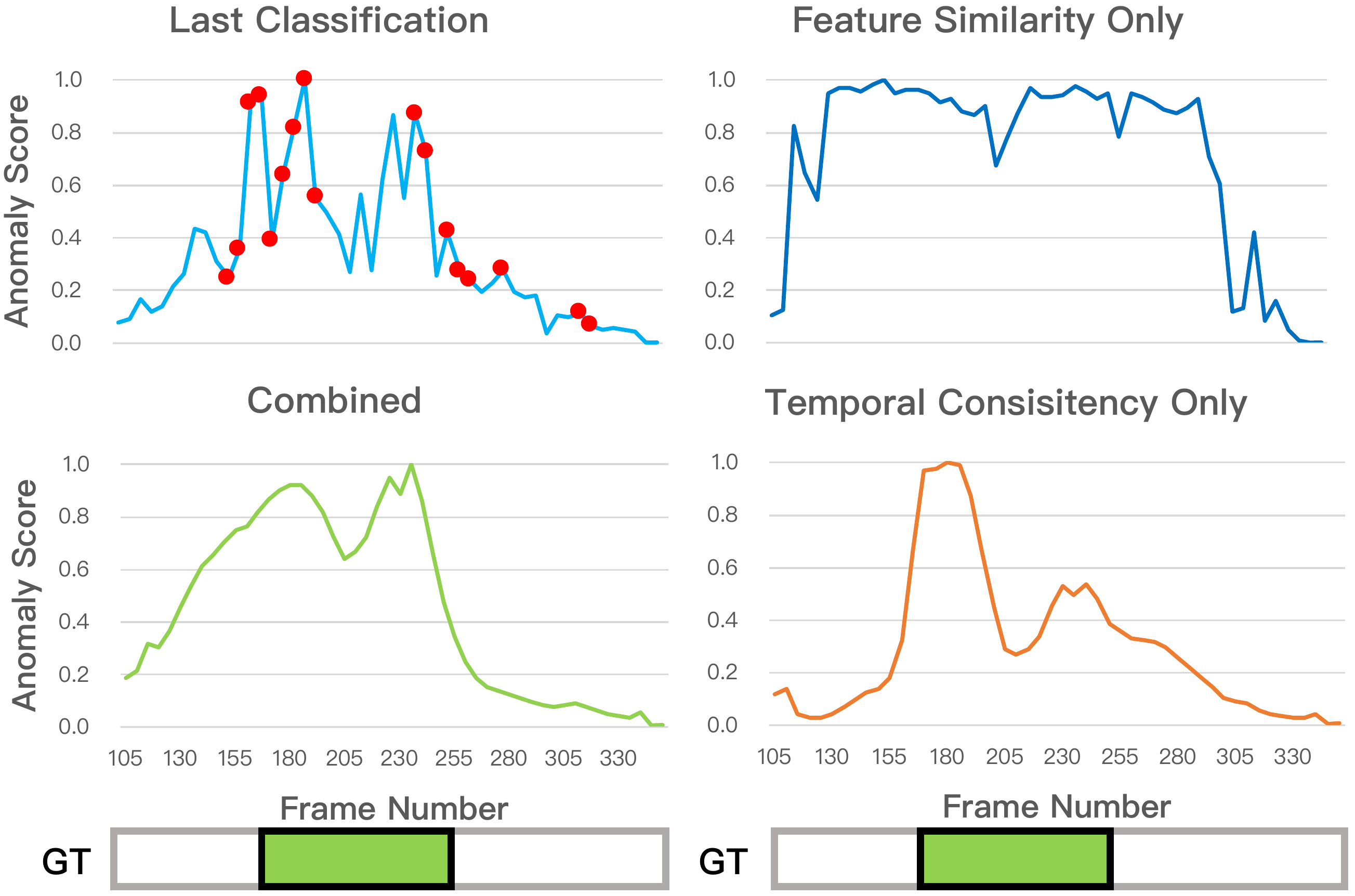}
\caption{\emph{Visualization of GCN outputs on ShanghaiTech w.r.t. the video ``05\_0021''.} The rough prediction at the upper left is from the optic flow branch, while the other three are snippet-wise labels cleaned by the GCN modules.}
\label{fig:SH_tech_vis}
\end{figure}

\textbf{Qualitative Analysis.}
Different from UCF-Crime, the training data in the new split of ShanghaiTech have temporal ground truths. Based on this, the working principle of our GCN can be intuitively understood. The anomaly event in Figure~\ref{fig:SH_tech_vis} is that a student jumps over the rail as shown in Figure~\ref{fig:vis_abvideo}. The temporal consistency module (at the upper right) is inclined to smooth the original high-confidence predictions (orange points at the upper left). Therefore, it correctly annotates the \(150^{th}-200^{th}\) frames with dense high-confidence predictions, but neglects the remaining ground truth for insufficient high-confidence inputs. The feature similarity module (at the lower right) tends to propagate information through a similar degree. It labels a long interval of snippets including the student's previous run-up and subsequent slow-down actions, possibly because they have the similar representation of ``a fast movement in the same direction'' on the optical flow. The entire GCN (at the lower left) combining these two modules can make more precise labels.

\subsection{Experiments on UCSD-Peds}

\begin{table}\centering
  \caption{\emph{Quantitative comparison on UCSD-Peds2}. Following the reviewer comment, we make more comparisons as shown in \textbf{Appendix}.}\label{tab:ucsd}
  \begin{tabular}{cc}
    \hline
    \textbf{Method}  &  \textbf{AUC (\%)}\\
    \hline\hline
    Adam~\cite{adam2008monitors} & 63.0 \\
    MDT~\cite{mahadevan2010anomaly} & 85.0 \\
    SRC~\cite{cong2011sparse} & 86.1  \\
    AMDN~\cite{xu2015deep} & 90.8 \\
    AL~\cite{he2017anomaly} & 90.1 \\
    \hline
    \textbf{Ours}\\
    TSN$^{Gray-scale}$ & 93.2 $\pm$ 2.3\\
    TSN$^{Optical Flow}$ & 92.8 $\pm$ 1.6\\
    \hline
  \end{tabular}
\end{table}

\begin{figure}[t]
    \captionsetup[subfigure]{position=bottom}
    \captionsetup[subfloat]{captionskip=0pt}
    \centering  
    \subfloat[RGB]{\animategraphics[loop,autoplay,poster=17,width=0.295\textwidth]{12}{figures/05_0021/rgb/img_}{050}{150}}
  	\subfloat[Flow-X]{\animategraphics[loop,autoplay,poster=17,width=0.295\textwidth]{12}{figures/05_0021/flow_x/flow_x_}{050}{150}}
  	\subfloat[Flow-Y]{\animategraphics[loop,autoplay,poster=17,width=0.295\textwidth]{12}{figures/05_0021/flow_y/flow_y_}{050}{150}}
    
    \caption{\emph{Partial video of ``05\_0021'' on ShanghaiTech.} \color{cyan}{Best viewed in Adobe Reader where (a)-(c) should play as videos.}}
    \label{fig:vis_abvideo}
%    \vspace{-15pt}
\end{figure}

In UCSD-Peds, some of the ground truths are only 4 frames, but the predictive unit of C3D reaches a length of 16 frames. Thus we conduct the experiments with TSN. To match the input dimension with the RGB branch, the original gray-scale frames are duplicated into the 3 primary-color channels. 

\begin{figure}[h]
  \centering
  \subfloat{
  	\label{subfig:ucsd_flow} 
  	\includegraphics[width=0.47\textwidth]{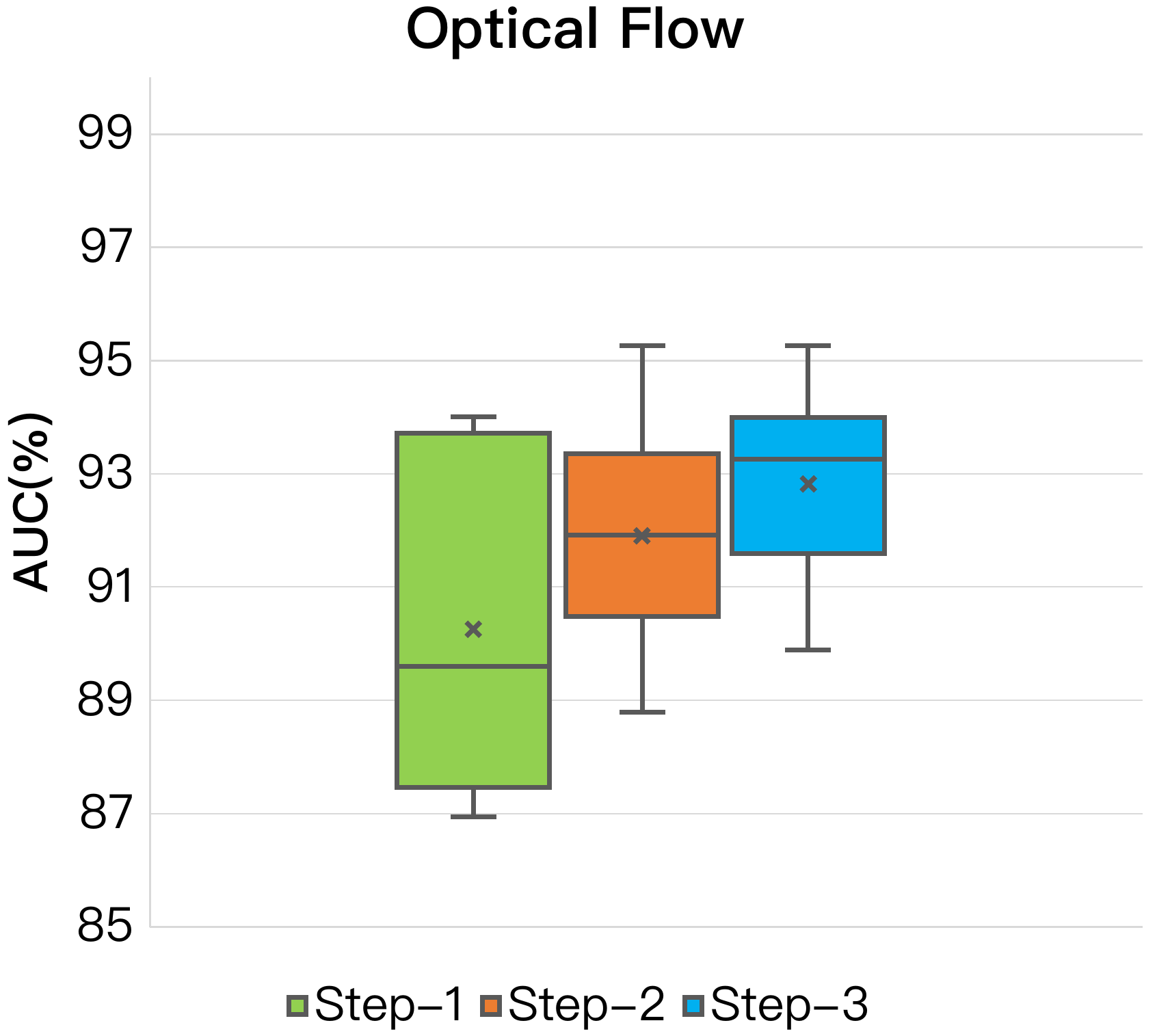}
  }
  \subfloat{
    \label{subfig:ucsd_rgb} 
    \includegraphics[width=0.47\textwidth]{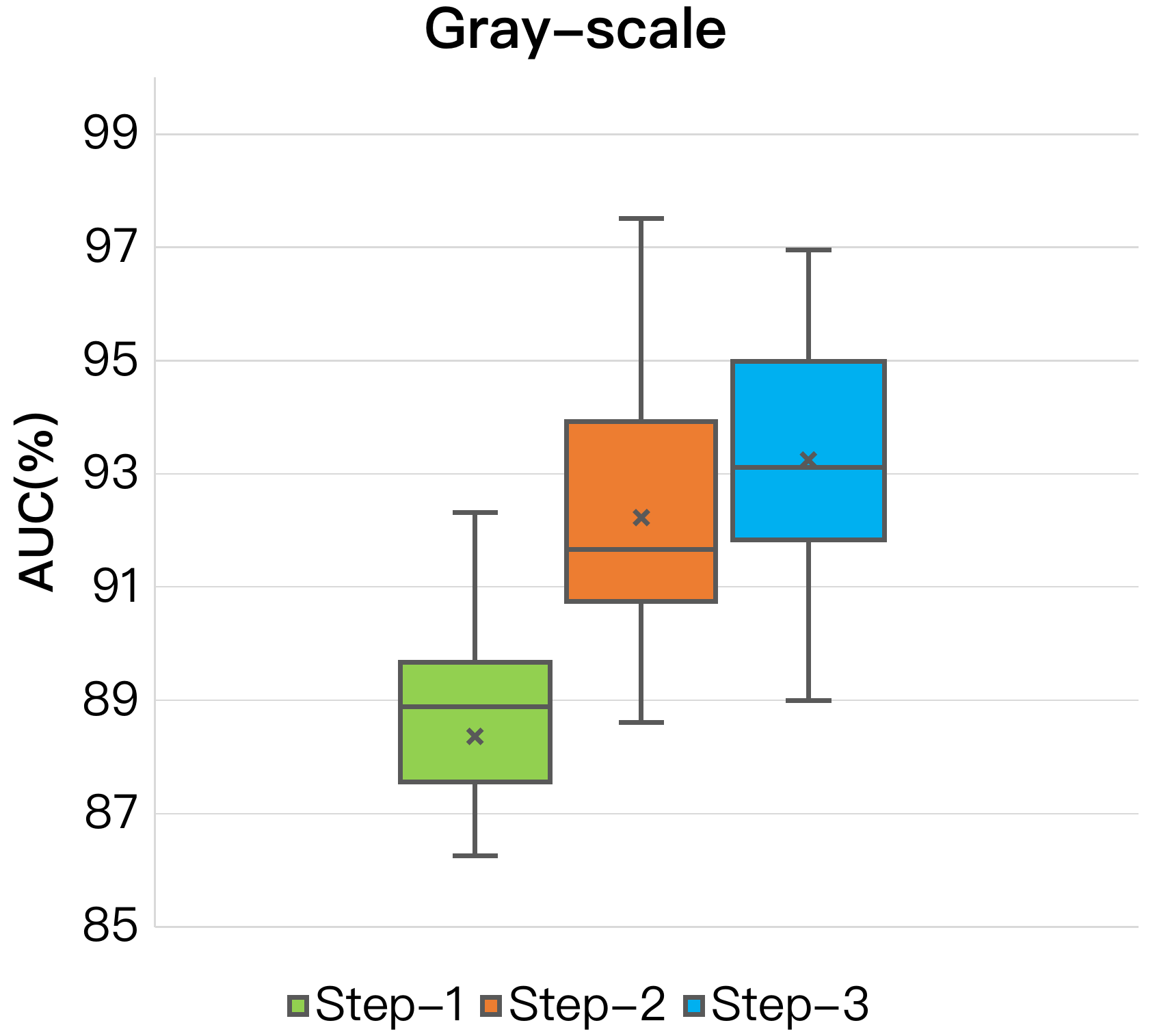}
    }
    
%   \vspace{-1.0em}
  \caption{\emph{Box-whisker plots of step-wise performance on UCSD-Peds2.}}
  \label{fig:ucsd} 
\end{figure}

\textbf{Step-wise results.}
After repeating experiments 10 times, we obtain the box plots in Figure~\ref{fig:ucsd}. The average results at the first step are good enough, so we start with feeding top 90\% high-confidence predictions into the GCN. We observe that the proposed method not only increases the detection performance, but also stabilize the predictions of the 10-time repeated experiments.

\textbf{Quantitative comparison.}
We report the ``mean value $\pm$ standard deviation'' of the AUC, and make comparisons with other methods under the same splitting protocol as in ~\cite{he2017anomaly}. Our approach outperform others with both the input modalities as shown in Table~\ref{tab:ucsd}.

\section{Conclusion}
In this paper, we address weakly supervised anomaly detection from a new perspective, by casting it as a supervised learning task under noise labels. In contrast to MIL formulation in previous works, such a perspective possesses distinct merits in two aspects: a) it directly inherits all the strengths of well-developed action classifiers; b) anomaly detection is accomplished by an integral end-to-end model with great convenience. Furthermore, we utilize a GCN to clean labels for training an action classifier. During the alternate optimization process, the GCN reduces noise via propagating anomaly information from high-confidence predictions to low-confidence ones. We validate the proposed detection model on 3 different-scale datasets with 2 types of action classification networks, where the superior performance proves its effectiveness and versatility.

\noindent\textbf{Acknowledgement.} This work was supported in part by the Project of National Engineering Laboratory-Shenzhen Division for Video Technology, in part by the National Natural Science Foundation of China and Guangdong Province Scientific Research on Big Data (No. U1611461), in part by Shenzhen Municipal Science and Technology Program (Grant JCYJ20170818141146428), and in part by National Natural Science Foundation of China (No. 61602014). We are grateful to the three anonymous reviewers for their valuable comments and suggestions. In addition, we would like to thank Jerry for English language editing.

\newpage
{\small
\bibliographystyle{ieee}
\bibliography{egbib}
}


\section*{Supplementary material}

# Graph Convolutional Label Noise Cleaner: Train a Plug-and-play Action Classifier for Anomaly Detection Supplementary Materials (Appendix)

Jia-Xing Zhong[1,2]   Nannan Li[3,1,2]   Weijie Kong[1,2]   Shan Liu[4]   Thomas H. Li[1]   Ge Li ✉[1,2]

[1]School of Electronic and Computer Engineering, Peking University   [2]Peng Cheng Laboratory
[3]Institute of Intelligent Video Audio Technology, Longgang Shenzhen   [4]Tencent America

jxzhong@pku.edu.cn    lnnsiat@gmail.com    weijie.kong@pku.edu.cn
shanl@tencent.com    tli@aiit.org.cn    geli@ece.pku.edu.cn

## 1. Processing Speed in the Test Phase

|  | Input Size (Pixel) | Speed (FPS) |
| --- | --- | --- |
| C3D | $112 \times 112$ | 123.08 |
| TSN-RGB | $224 \times 224$ | 30.96 |
| TSN-Optical Flow | $224 \times 224$ | 150.15 |

Table 1: *Testing speed (FPS) of our models on a Titan-XP GPU.* Note that the reported result includes all pre-processing operations, such as resizing, 10-crop oversampling, zero-centering, *etc*.

Our approach of directly utilizing action classifiers for anomaly detection has great computational efficiency. As shown in Table 1, we report the frame per second (FPS) performance of the two types of action classifiers. Although the time-consuming pre-processes (*e.g.*, 10-crop oversampling) are taken into consideration, the three action classifiers still have the real-time or even the super real-time performance.

## 2. Implementation of Label Noise Cleaner

At the first cleaning step, we select the 30% and 60% highest-confidence snippets as $H$ for two-stream and C3D networks respectively if not specified, and increase the cardinality of $H$ by 30% at each step. To learn an unbiased model, we also include normal videos in training data. To generate the label assignments of action classifiers, we concentrate the output probability into a single anomaly category with a min-max normalization. The output dimensions of the first two fully connected layers are 512 and 128 respectively, at the 60% dropout [10] rate. Both the graph modules have two convolutional layers: a 32-unit hidden layer activated by ReLu and the last 1-unit output layer. Due to the limited memory of GPUs, we at most sample 1,600 high-confidence snippets with not more than 8 neighbours respectively in a video. We implement our noise cleaner upon Pytorch [8] with the following hyper-parameters: $base\_learning\_rate = 0.0001$, $momentum = 0.9$ and $weight\_decay = 0.0005$. In preliminary experiments, we observe that three iterations are sufficient in most cases. Therefore we repeat the alternate optimization until the $3^{rd}$ step and compare the last (not always the best) results with other methods.

## 3. More Comparisons on UCSD-Peds

Several unary-classification works in 2018 also conduct experiments on *UCSD-Peds*. As shown in Table 3 and the main body of our paper, their default implementations are not directly comparable with ours because of different data splits. *For some open-source works, we hereby reproduce experiments on the data split in [1] as ours,* while the results in their original papers are also provided within square parentheses "[]" for reference as reported in Table 2. Since *UCF-Crime* is released at Github on June $10^{th}$ 2018 lately, except the official reference [11] and its comparisons, neither public reporting of results nor source codes can be found, and we hope that our work can fill in the blanks.

## 4. Vectorized Feature Similarity Module

Following the main body of our paper, we denote the feature similarity graph as $F = (V, E, \mathbf{X})$, where $V$ is the vertex set, $E$ is the edge set, and $\mathbf{X}$ is the attribute of vertexes. In particular, $V$ is a video, $E$ describes the feature similarity amongst snippets, and $\mathbf{X} \in \mathbb{R}^{\mathbf{N} \times \mathbf{d}}$ represents the $d$-dimensional feature of these $N$ snippets. The adjacency matrix $\mathbf{A} \in \mathbb{R}^{N \times N}$ of $F$ is defined as:

$$\mathbf{A}_{(i,j)} = \exp(\mathbf{X}_i \cdot \mathbf{X}_j - max(\mathbf{X}_i \cdot \mathbf{X})), \qquad (1)$$

1

| Method | Publication | AUC (%) |
|---|---|---|
| **Unary-classification Paradigm** | | |
| ... | ... | ... |
| TCP [9] | WACV 2018 | No source codes [88.4] |
| Frame Prediction [6] | CVPR 2018 | **92.6 ± 1.1 [95.4]** |
| C2ST [7] | BMVC 2018 | 81.4 ± 2.8 [87.5] |
| **Binary-classification Paradigm** | | |
| AL [1] | J-Mult. Nov. 2018 | 90.1 |
| Ours-TSN$^{Gray-scale}$ | – | **93.2 ± 2.3** |
| Ours-TSN$^{OpticalFlow}$ | – | 92.8 ± 1.6 |

Table 2: *Comparison on UCSD-Peds in 2018.* The results of their original papers under data split [5] are reported within "[]".

| Splitting Approach | Train | | Test |
|---|---|---|---|
| | Normal | Abnormal | |
| Following [5] | 16 | 0 | 12 |
| Following [1] | 4 | 6 | 18 |

Table 3: *Difference in splitting UCSD-Peds.* The random selection is repeated 10 times in [1].

where the element $\mathbf{A}_{(i,j)}$ measures the feature similarly between the $i^{th}$ and $j^{th}$ snippets. Here is an equivalent vectorization of Equation 1:

$$\mathbf{A} = \exp(\mathbf{X}\mathbf{X}^T - torch.max(\mathbf{X}\mathbf{X}^T, dim=1)), \quad (2)$$

where the $torch.max$ function takes the maximum value over dimension 1.

The nearby vertexes are driven to have the same anomaly label via the graph-Laplacian operation approximated with a renormalization trick [3]:

$$\widehat{\mathbf{A}} = \widetilde{\mathbf{D}}^{-\frac{1}{2}} \widetilde{\mathbf{A}} \widetilde{\mathbf{D}}^{-\frac{1}{2}}, \quad (3)$$

where the self-loop adjacency matrix $\widetilde{\mathbf{A}} = \mathbf{A} + \mathbf{I_n}$, and $\mathbf{I_n} \in \mathbb{R}^{N \times N}$ is the identity matrix; $\widetilde{\mathbf{D}}$ is the corresponding degree matrix:

$$\widetilde{\mathbf{D}}_{(i,i)} = \sum_j \widetilde{\mathbf{A}}_{(i,j)}. \quad (4)$$

The vectorization of Equation 4 is implemented with the vectorized summation and the broadcasting diagonal functions of Pytorch:

$$\widetilde{\mathbf{D}} = torch.diag(torch.sum(\widetilde{\mathbf{A}}, dim=1)). \quad (5)$$

Finally, the output $\mathbf{H}$ of a feature similarity graph module layer is computed as:

$$\mathbf{H} = \sigma(\widehat{\mathbf{A}}\mathbf{X}\mathbf{W}), \quad (6)$$

where $\mathbf{W}$ is a trainable parametric matrix, and $\sigma$ is an activation function.

Since the whole computational procedure is differentiable, our feature similarity graph module can be trained in an end-to-end fashion. Therefore, neural networks are capable of seamlessly incorporating the single or multiple stacked modules. The temporal similarity module can be also rewritten as its corresponding vectorized implementation in a similar manner.

## 5. Details of Indirectly Supervised Loss Term

Our indirectly supervised term of the loss function can be viewed as a temporal ensembling strategy [4]. The pseudo code is shown in Algorithm 1. In practice, we set $\gamma$ as 0.5 in all of the experiments. Since we have already obtained a set of rough predictions from the action classifier, the "cool start" initialization and the bias correction of the original temporal ensembling method [4] are not required as illustrated on the $1^{st}$ and the $8^{th}$ statements.

## 6. Reorganization of ShanghaiTech

| | Training Set | Test Set | Total |
|---|---|---|---|
| **Normal Videos** | 175 | 155 | 330 |
| **Anomaly Videos** | 63 | 44 | 107 |
| **Total** | 238 | 199 | 437 |

Table 4: *The number of videos on our reorganized ShanghaiTech.*

In total, there are 437 videos on ShanghaiTech. As shown in Table 4, we split the data into two subsets: the training set is made up of 238 videos, and the testing set contains 199 videos. In each scene, the numbers of normal and anomaly videos w.r.t. the two subsets are depicted in Figure 1 and Figure 2, respectively. The new

data split is available at https://github.com/jx-zhong-for-academic-purpose/GCN-Anomaly-Detection.

## 7. Discuss the Formulation

Following the reviewer's suggestion, we discuss our noisy-labeled problem formulation and the EM-like optimization mechanism under this formulation in more detail.

### 7.1. Concept: MIL vs Noisy-labeled Learning

Conceptually, the two formulations mainly differ in their *emphases*. Given a positive bag $Y = 1$, the MIL usually focuses on *positive instances* $y_i = 1$, whereas the noisy-labeled training pays attention to *noisy labels* $y_i = 0$ and the remaining ones are $y_i = 1$. The two conceptions are complementary and have transformational relations.

### 7.2. Practice: EM-like MIL vs Ours

Practically, in terms of *selection criteria on "seed examples"*, the EM-like MIL focuses on the *most-likely positive instances*, while our noisy-labeled optimization prefers the *most-likely reliable predictions*. Take the three MIL models the reviewer mentioned for examples. If the 10-crop prediction of a snippet within an anomalous video is {0.2, 0.2, ..., 0.2}, He *et al.* [1] will not update their "anchor dictionary" with it for its low anomaly score *(mean value=0.2)*, Hou *et al.* [2] will exclude it because it is "non-discriminative" (without "the same label" as the corresponding video), Zhang *et al.* [12] will neglect it since their E-step is to seek the most "responsible" instance to the bag annotation, but we will select it to supervise our GCN because it is highly certain and noiseless *(predictive variance=0)*.

### 7.3. Terminology: EM-like vs EM-based

As pointed out in the main body of this paper, our updating method is "EM-like" instead of "EM-based". The resemblance between our optimization mechanism and the EM-based approach is that they both alternately repeat update-and-fix processes. However, our method is not "EM-based" since we do not explicitly estimate mathematical expectation in the training process.

**Algorithm 1** *Indirectly Supervised Loss Term.*

Note that the practical computational processes are incrementally implemented, while in this pseudo code all of them are calculated from the $1^{st}$ epoch for clarity.

**Input:**
  $V = \{v_i\}_{i=1}^{N}$: a video with $N$ snippets
  $\widetilde{Y} = \{\widetilde{y}_i\}_{i=1}^{N}$: the rough snippet-wise anomaly probabilities from the last action classifier
  $p_\theta(v_i)$: the GCN predictions of video clips $v_i$ with trainable parameters $\theta$
  $\alpha(v_i)$: the stochastic augmentation (such as dropout and random cropping) function of input snippets $v_i$
  $\gamma$: a hyper-parametric discount factor within the range of $(0, 1)$

**Output:**
  $\mathcal{L}_I^j$: the indirectly supervised loss at the $j^{th}$ epoch

1: Initialize the smooth target $\overline{p}_{i \in 1,2,..,N} = \widetilde{y}_{i \in 1,2,..,N}$
2: **repeat**
3:   Initialize the epoch counter $j = 0$
4:   **for** each video $V$ in the training set **do**
5:     Obtain the GCN predictions of augmented snippets: $p_i = p_\theta(\alpha(v_i))$
6:     Compute the loss under indirect supervision: $\mathcal{L}_I^j = \frac{1}{N}\sum_{i=1}^{N}|p_i - \overline{p}_i|$
7:     Optimize the parameters $\theta$ of the GCN
8:     Update the smooth target: $\overline{p}_{i \in 1,2,..,N} = \gamma \overline{p}_{i \in 1,2,..,N} + (1-\gamma)p_{i \in 1,2,..,N}$
9:   Update the epoch counter: $j = j + 1$
10: **until** j == current epoch number

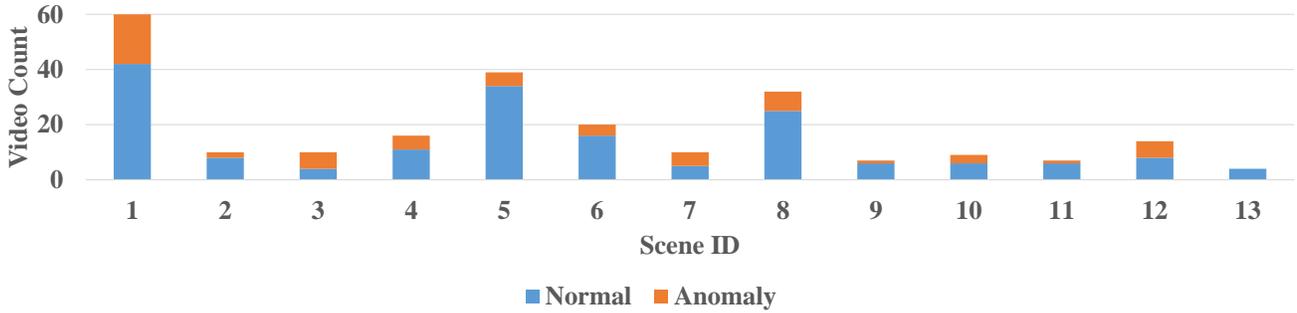

Figure 1: *Training set on the reorganization of ShanghaiTech.*

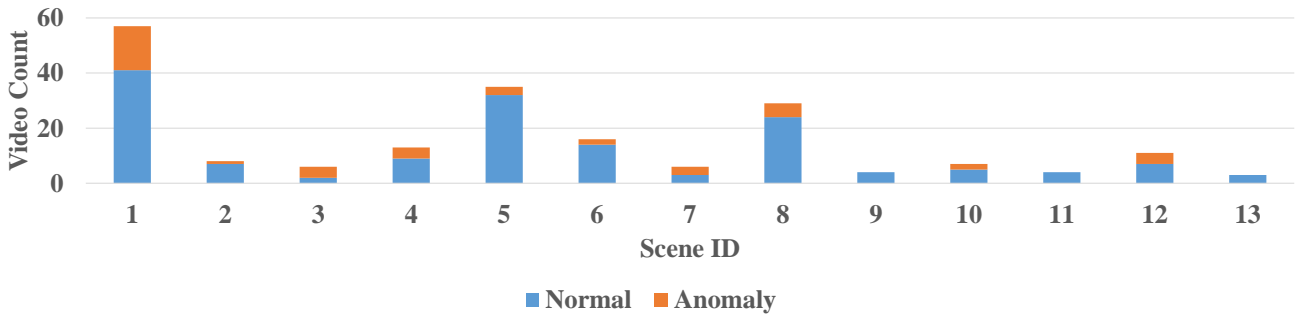

Figure 2: *Testing set on the reorganization of ShanghaiTech.*


\end{document}